\documentclass[11pt]{article}

\usepackage[preprint]{acl}

\usepackage{times}
\usepackage{latexsym}

\usepackage[T1]{fontenc}

\usepackage[utf8]{inputenc}

\usepackage{microtype}

\usepackage{inconsolata}

\usepackage{graphicx}

\usepackage{amsmath}
\usepackage{amsfonts}
\usepackage{booktabs}
\usepackage{subcaption}
\usepackage{tcolorbox}

\usepackage{soul}
\sethlcolor{yellow!30}

\usepackage{tikz}
\usetikzlibrary{arrows.meta, calc}
\definecolor{arrowgreen}{RGB}{0,140,70}
\definecolor{circlepurple}{RGB}{160,50,160}
\definecolor{overlapcolor}{RGB}{180,50,50}

\usepackage{etoolbox}
\newtoggle{release}
\toggletrue{release}

\iftoggle{release}{
\newcommand{\alex}[1]{}
\newcommand{\eugene}[1]{}
\newcommand{\atsu}[1]{}
}
{
\newcommand{\alex}[1]{{\color{blue} A: #1}}
\newcommand{\eugene}[1]{{\color{red} G: #1}}
\newcommand{\atsu}[1]{{\color{magenta}Y: #1}}
}

\title{Multi-Faceted Interactivity Alignment in Full-Duplex Speech Models}

\author{
  Atsumoto Ohashi$^{1}$,\hspace{0.5em}
  Neil Zeghidour$^{2}$,\hspace{0.5em}
  Alexandre D\'{e}fossez$^{1,2}$\thanks{These authors jointly supervised this work.},\hspace{0.5em}
  Eugene Kharitonov$^{2}$\footnotemark[1] \\[0.5em]
  $^{1}$Kyutai, Paris, France \quad $^{2}$Gradium, Paris, France \\[0.3em]
  \texttt{\{atsumoto.ohashi, alex\}@kyutai.org, eugene@gradium.ai}
}

\begin{document}
\maketitle
\begin{abstract}
Full-duplex spoken dialogue models can listen and speak simultaneously, making them a promising architecture for natural conversation. However, current models are trained solely with supervised learning through token-level likelihood maximization, which does not directly optimize interaction-level behaviors, causing interactivity issues such as excessive silence and ill-timed turn-taking. Recent work has applied reinforcement learning (RL) to improve interactivity, but existing methods address only a limited set of interactive behaviors in their rewards. In this work, we propose a post-training alignment method that comprehensively improves the interactivity of full-duplex spoken dialogue models through RL. We address the four canonical axes of interactivity: pause handling, turn-taking, backchanneling, and user interruption. For each axis, we extract short audio segments from human conversation corpora and optimize the model with axis-specific reward functions. An extra LLM-based reward for response quality prevents semantic degradation. We apply our method to two open-source models, Moshi and PersonaPlex, demonstrating consistent improvements in interactivity on both offline evaluation with pre-recorded audio and real-time multi-turn dialogue evaluation.\footnote{The checkpoints of the models and audio samples are available at \url{https://huggingface.co/kyutai/moshika-rl-seamless} and \url{https://huggingface.co/kyutai/personaplex-rl-seamless}.}
\end{abstract}

\section{Introduction}
Full-duplex spoken dialogue models are a promising architecture for reproducing human-like conversational dynamics in spoken dialogue systems~\citep{ji2024wavchat, arora2026landscape}. Conventional turn-based spoken dialogue models~\citep{zhang2023speechgpt, fang2024llamaomni, xu2025qwen25omni, wu2025stepaudio} begin generating a response only after the user's entire utterance has been received, and the user must wait for the system to finish speaking before taking turns. In contrast, full-duplex models~\citep{defossez2024moshi, zhang2025omniflatten, yu2025salmonnomni} are designed to process both the user's and the system's speech streams in parallel. This eliminates the need for explicit turn boundary detection via external voice activity detection (VAD) modules, and enables the model to implicitly learn smooth turn-taking, backchanneling, and overlap handling within its internal representations.

Yet, full-duplex models still face challenges in interactivity in real-time conversations. Previous studies~\citep{arora2024talking, lin2025fullduplexbench} have revealed issues such as excessive silence, badly timed turn-taking, and a lack of backchannels. One possible cause is the inherent limitation of supervised learning. Interaction-level behaviors, such as the timing and duration of utterances, are difficult to optimize directly through token-level likelihood maximization. Furthermore, the mismatch between the input distributions at training and inference time, such as exposure bias, prevents models from behaving robustly in real conversations.

Prior works have explored using reinforcement learning (RL) to improve the interactivity of full-duplex models~\citep{yu2025salmonnomni, chen2025reinforcement}, but covered only a subset of conversational dynamics such as handling users' barge-ins and backchanneling, failing to comprehensively address all axes of interactivity. Furthermore, each study targets a single model, leaving open the question of whether RL-based interactivity optimization generalizes across different full-duplex systems. A common practical challenge is also that optimizing timing-related rewards alone can degrade the semantic quality of generated responses~\citep{hsiao2026aspirin}, yet no systematic solution has been established.

In this work, we propose a post-training alignment method that comprehensively improves the interactivity of full-duplex speech models. We target the four core axes of full-duplex interactivity: pause handling, turn-taking, backchanneling, and user interruption. We automatically extract short audio segments from real human conversations exhibiting such behaviors, and optimize the model on them using group relative policy optimization (GRPO)~\citep{shao2024deepseekmath}. Furthermore, an LLM Judge reward for semantic quality~\citep{lin2025alignslm, arora2026optimizing} preserves response content during RL. We apply the proposed method to two different open-source models, Moshi~\citep{defossez2024moshi} and PersonaPlex~\citep{roy2026personaplex}, and demonstrate consistent improvements across all metrics on Full-Duplex-Bench v1~\citep{lin2025fullduplexbench}. Notably, although training is performed on short, extracted segments, we also demonstrate that the improvements generalize to real-time multi-turn dialogues through the evaluation on Full-Duplex-Bench v2~\citep{lin2026fullduplexbenchv2}. A subjective evaluation further confirms that human raters consistently prefer the RL-trained models.

\section{Related Work}

\subsection{Full-Duplex Spoken Dialogue Models}
Full-duplex models can be broadly categorized into two types: cascaded and end-to-end. Cascaded approaches achieve full-duplex dialogue by augmenting text-based LLMs with external VAD and turn-control modules~\citep{wang2024fullduplex, zhang2024turnbased, fu2025vita, wang2025freezeomni, liao2025flexduo}, but they suffer from inter-module latency and loss of paralinguistic information. End-to-end approaches, in contrast, process both the user's and the model's audio streams within a unified model~\citep{nguyen2023generative, hu2025efficient, wang2025ntpp}. Various strategies exist for handling the two streams: processing them fully in parallel~\citep{defossez2024moshi, yao2026flmaudio, shi2025voila}, interleaving them in short alternating chunks~\citep{veluri2024turnbased, zhang2025omniflatten}, or explicitly controlling listen/speak states~\citep{yu2025salmonnomni, chen2025minmo}. Most existing models are trained solely with supervised learning, and are not optimized for interaction-level properties. In this work, we propose a post-training method based on RL to unlock the full potential of end-to-end architectures for improved interactivity.


\subsection{Reinforcement Learning for Spoken Language Models}
Inspired by the success of alignment techniques~\citep{stiennon2020learning, ouyang2022training, rafailov2023direct} in text language modeling, these methods have been extended to speech for semantic coherence~\citep{lin2025alignslm}, speech quality~\citep{zhang2024speechalign}, question answering~\citep{chen2026wavalign}, reasoning~\citep{wu2025stepaudio}, and paralinguistic processing~\citep{yang2025paras2s}.

For full-duplex dialogue models specifically, alignment methods have been applied not only to improve semantic quality~\citep{arora2026optimizing, wu2025aligning} but also to optimize dialogue dynamics~\citep{yu2025salmonnomni, chen2025reinforcement}. SALMONN-omni~\citep{yu2025salmonnomni} applies direct preference optimization (DPO)~\citep{rafailov2023direct} to improve the model's ability to handle users' barge-ins and backchannels. ORISE~\citep{chen2025reinforcement} replaces preference-based methods with online RL based on REINFORCE~\citep{williams1992simple}, achieving a better balance between barge-in and backchannel handling in noisy environments. 
The most closely related work is the concurrent ASPIRin~\citep{hsiao2026aspirin}, which also applies GRPO to Moshi to optimize response timing. However, these prior and concurrent works incorporate only a subset of interactivity aspects into their reward functions, and report that other aspects, such as response latency or semantic quality, either fail to improve or even degrade~\citep{hsiao2026aspirin}. In this work, we comprehensively address four axes of interactivity, together with LLM-based evaluation of response content quality. We demonstrate that our approach improves performance across all aspects, including those that degraded in prior work.

\begin{figure*}
\centering
\includegraphics[width=1\linewidth]{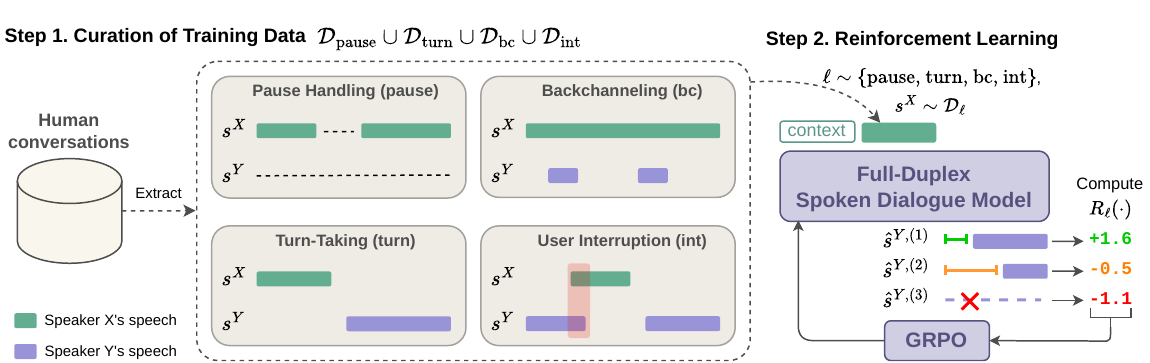}
\caption{Overview of the proposed method. We first extract segments related to each interactivity axis $\ell \in \{\mathrm{pause}, \mathrm{turn}, \mathrm{bc}, \mathrm{int}\}$ from human conversation datasets to construct the training data $\mathcal{D}_\ell$ . For each segment (and optionally its dialogue context) sampled from $\mathcal{D}_\ell$, the full-duplex spoken dialogue model generates multiple outputs, which are scored by axis-specific reward functions $R_\ell$ and used to optimize the model via GRPO.}
\label{fig:method}
\end{figure*}

\subsection{Benchmarks for Full-Duplex Spoken Dialogue Models}
Earlier work relied on corpus-level statistics such as distributions of utterance and silence durations~\citep{nguyen2023generative} to assess full-duplex dialogue models. Benchmarks like Full-Duplex-Bench v1~\citep{lin2025fullduplexbench} have since been developed to enable more fine-grained evaluation. It feeds pre-recorded static audio into dialogue models and evaluates the resulting responses~\citep{peng2025fdbench, ge2025flexi, chang2026gametime}. More recently, benchmarks have been constructed for dynamic evaluation in multi-turn dialogues, going beyond static assessment~\citep{arora2024talking, zhang2026mtrduplexbench}. Full-Duplex-Bench v2~\citep{lin2026fullduplexbenchv2} has dialogue models interact in real time with an automated conversational partner and evaluates their abilities using LLM Judge. In this work, we leverage the four evaluation axes of Full-Duplex-Bench v1 for RL reward design, and use Full-Duplex-Bench v2 to verify generalization to multi-turn settings.

\section{Method}
\label{sec:method}

We propose a post-training alignment method that improves the interactivity of full-duplex spoken dialogue models through RL. We target four core axes of interactivity: \emph{pause handling} (remaining silent during user hesitations), \emph{turn-taking} (responding promptly when the user yields the floor), \emph{backchanneling} (producing short feedback cues while the user speaks), and \emph{user interruption} (yielding and responding when the user barges in). These four axes have been established as a standard and comprehensive characterization of full-duplex interactivity~\citep{lin2025fullduplexbench}, and we adopt them as a necessary set for our study.\footnote{While our method can readily accommodate additional axes, exploring them is beyond the scope of this work.} Figure~\ref{fig:method} illustrates the overview of our method. Our key idea is to extract short audio segments from real human conversations, each exemplifying one of the four axes, and optimize the model on them using GRPO~\citep{shao2024deepseekmath} with axis-specific rewards.

Below, we first describe the full-duplex modeling framework assumed in this work (Section~\ref{ssec:method_fdsdm}), followed by our RL pipeline (Section~\ref{ssec:method_optimization}), and its two key components: segment extraction (Section~\ref{ssec:method_data_curation}) and reward design (Section~\ref{ssec:method_reward}).

\subsection{Full-Duplex Spoken Dialogue Modeling}
\label{ssec:method_fdsdm}

We consider full-duplex spoken dialogue models based on discrete audio tokens and autoregressive language modeling~\citep{defossez2024moshi, veluri2024turnbased}. Given a two-channel dialogue between speakers $X$ and $Y$, a speech tokenizer $E$ maps each speaker's waveform into a sequence of discrete tokens from a vocabulary $\mathcal{V}$: $[x_{1:N};\, y_{1:N}] = E([s^X;\, s^Y])$. The model, parameterized by $\theta$, learns to autoregressively predict speaker $Y$'s tokens conditioned on speaker $X$'s input stream and its own preceding outputs:
\begin{equation*}
\resizebox{1.0\linewidth}{!}{$\displaystyle
\pi_\theta(y_{1:N}, w_{1:N} \mid x_{1:N})
= \prod_{n=1}^{N} \pi_\theta(y_n, w_n \mid x_{\leq n}, y_{<n}, w_{<n})\;.
$}
\end{equation*}
Here, we assume that a parallel text token stream $w_{1:N}$ is jointly predicted, which is a common practice used to guide the semantic content of the generated speech~\citep{chen2025minmo, yao2026flmaudio} and also to implicitly control \emph{when} the model speaks~\citep{defossez2024moshi, shi2025voila}. After the supervised learning on dialogue datasets, the model $\pi_\theta$ can serve as spoken dialogue systems.

\subsection{Reinforcement Learning Pipeline}
\label{ssec:method_optimization}

At each training step during RL of the pre-trained model $\pi_\theta$, for each sample in the batch, we first sample an interactivity axis $\ell \in \{\mathrm{pause}, \mathrm{turn}, \mathrm{bc}, \mathrm{int}\}$, corresponding to pause handling, turn-taking, backchanneling, and user interruption, respectively. We then draw a segment from the axis-specific training set $[s^X; s^Y] \sim \mathcal{D}_\ell$, which consists of short audio clips extracted from human conversation corpora (see Section~\ref{ssec:method_data_curation} for extraction details). From the speaker $X$'s audio in the segment, which is encoded into tokens $x_{1:N}$, the current policy $\pi_\theta$ generates $G$ completions:
\begin{equation*}
\{(\hat{y}^{(g)}_{1:N},\; \hat{w}^{(g)}_{1:N})\}_{g=1}^{G}
\;\sim\; \pi_\theta(\cdot \mid x_{1:N})\;.
\end{equation*}
Each completion is decoded into a waveform $\hat{s}^{Y,(g)} = D(\hat{y}^{(g)}_{1:N})$ and scored by an axis-specific reward function $r^{(g)} = R_\ell(\hat{s}^{Y,(g)})$ (see Section~\ref{ssec:method_reward}).

We normalize the rewards across the $G$ samples to get an advantage estimate for each completion:
\begin{equation*}
\hat{A}^{(g)}
\;=\;
\frac{r^{(g)} - \mathrm{mean} ( \{r^{(g')} \}_{g'=1}^G)}{ \mathrm{std}(\{r^{(g')}\}_{g'=1}^G)}\;.
\end{equation*}
We then minimize a clipped surrogate loss~\citep{schulman2017proximal} augmented with a KL penalty against the frozen reference policy $\pi_{\mathrm{ref}}$ (a copy of the model before RL training):

\begin{equation*}
\resizebox{1.0\linewidth}{!}{$\displaystyle
\begin{aligned}
\mathcal{L}(\theta)
&= -\frac{1}{G}\sum_{g=1}^{G} \frac{1}{N}\sum_{n=1}^{N}
\Bigl[
\min\!\bigl(\rho_n^{(g)}\,\hat{A}^{(g)},\\
&\qquad
\mathrm{clip}(\rho_n^{(g)},\, 1{-}\epsilon,\, 1{+}\epsilon)\,\hat{A}^{(g)}\bigr)
- \beta\,\mathrm{KL}\bigl[\pi_\theta \| \pi_{\mathrm{ref}}\bigr]_n
\Bigr]\;,
\end{aligned}
$}
\end{equation*}
where $\rho_n = \pi_\theta(\hat{w}_n \mid \cdot) / \pi_{\theta_{\mathrm{old}}}(\hat{w}_n \mid \cdot)$ is the importance sampling ratio between the current and the sampling-time policies, and $\mathrm{KL}[\cdot\|\cdot]_n$ is the exact KL divergence at position $n$. Note that because interactivity and the speech content are primarily controlled by text tokens~\citep{hsiao2026aspirin, chen2026wavalign}, $\rho_n$ and the objective are computed using only the probability of the text tokens $\hat{w}$, excluding the audio tokens $\hat{y}$.

To help the model generalize beyond the short segment boundaries, we prepend a context, which is the recording immediately preceding the segment, to the input audio. The context length is drawn randomly during training (details in Section~\ref{ssec:exp_model_train}). The loss is computed only over the segment itself; the context window is masked out.

\subsection{Training Data Curation}
\label{ssec:method_data_curation}

We construct the training set $\mathcal{D}_\ell$ for each interactivity axis $\ell \in \{\mathrm{pause}, \mathrm{turn}, \mathrm{bc}, \mathrm{int}\}$ from corpora of two-party human conversations in which each speaker is recorded on a separate channel. One channel serves as the user input ($X$) and the other as the target speaker ($Y$) whose behavior the model learns to reproduce. The extraction proceeds in two stages: utterance annotation and event-driven segment identification.

\paragraph{Utterance annotation}
We first annotate each recording using a VAD model, which produces a sequence of labeled intervals classified as inter-pausal units (IPUs) or silences for each speaker. We then group consecutive IPUs from the same speaker into \emph{utterances} by placing an utterance boundary at any silence longer than $1.0\;\text{s}$. A silence of at most $1.0\;\text{s}$ that falls within an utterance is called a \emph{pause}. Let $U_1, U_2, \dots, U_M$ denote the sequence of all utterances from both speakers, sorted by their start time. Each utterance $U_k$ is associated with a speaker $\mathrm{spk}(U_k)\in\{X, Y\}$, a start time $t_\mathrm{start}(U_k)$, and an end time $t_\mathrm{end}(U_k)$. We write $\mathrm{dur}(U_k) = t_\mathrm{end}(U_k) - t_\mathrm{start}(U_k)$ for the duration.

\paragraph{Event-driven segment identification}
From the utterance sequence, we identify \emph{segments} that exemplify each interactivity axis. Each segment defines a time window over the stereo recording, together with a label indicating its axis. The extraction criteria for each axis are as follows:

\begin{description}
\itemsep0em

\item[Pause Handling $\mathcal{D}_\mathrm{pause}$]
A single utterance $U_k$ with $\mathrm{spk}(U_k) = X$ satisfying:
(i)~$\mathrm{dur}(U_k) \geq \tau_{\min}$;
(ii)~$U_k$ contains at least one internal pause; and
(iii)~speaker $Y$ produces no speech.
This captures moments where $X$ hesitates but does not yield the floor.

\item[Turn-Taking $\mathcal{D}_\mathrm{turn}$]
A consecutive pair $(U_k,$ $U_{k+1})$ with $\mathrm{spk}(U_k) = X$ and $\mathrm{spk}(U_{k+1}) = Y$, where
(i)~both utterances satisfy the minimum duration $\tau_{\min}$; and
(ii)~the gap $t_\mathrm{start}(U_{k+1}) - t_\mathrm{end}(U_k) \leq 0.4\;\text{s}$~\citep{lin2025fullduplexbench, heldner2010pauses}.
This captures smooth turn transitions where $Y$ responds promptly after $X$ yields the floor.

\item[Backchanneling $\mathcal{D}_\mathrm{bc}$]
An utterance $U_k$ with $\mathrm{spk}(U_k)$ $= X$ satisfying:
(i)~$\mathrm{dur}(U_k) \geq \tau_{\min}$;
(ii)~speaker $Y$ says only short utterances (${\leq} \,1\,\text{s}$) in $U_k$~\citep{ekstedt2022voice}.

\item[User Interruption $\mathcal{D}_\mathrm{int}$]
A four-utterance sequence $(U_k, U_{k+1}, U_{k+2}, U_{k+3})$ with the speaker pattern $X \to Y \to X \to Y$:
$U_k$ is an initial utterance by $X$;
$U_{k+1}$ is $Y$'s response that gets interrupted;
$U_{k+2}$ is $X$'s interrupting utterance, which begins before $U_{k+1}$ ends; and
$U_{k+3}$ is the post-interruption response.
All four utterances must satisfy $\tau_{\min}$.
\end{description}

Rather than relying on TTS-synthesized dialogues or artificial noise augmentation~\citep{chen2025reinforcement}, our training data from human conversations can broadly capture real-world conversational dynamics and noise without artificial bias.

\subsection{Reward Design}
\label{ssec:method_reward}


We design a dedicated reward function $R_\ell$ for each interactivity axis $\ell \in \{\mathrm{pause}, \mathrm{turn}, \mathrm{bc}, \mathrm{int}\}$ to compute the reward of the generated audio $\hat{s}^{Y,(g)}$. We first apply VAD to $s^X$ and $\hat{s}^{Y,(g)}$ to obtain sequences of speech intervals and utterance annotations as in Section~\ref{ssec:method_data_curation}.

\begin{description}
\itemsep0em
\item[Pause handling $R_\mathrm{pause}$]
The model should remain silent throughout the segment, including during intra-utterance pauses. We assign a binary reward $R_\mathrm{pause} = -1$ if the generated audio contains any speech interval longer than $1\;\text{s}$, and $R_\mathrm{pause} = 0$ otherwise.

\item[Turn-Taking $R_\mathrm{turn}$]
The model should begin speaking promptly after speaker $X$ finishes. Let $d$ be the response delay between the end time of $X$'s turn and the start time of the first utterance ($>1\;\text{s}$) in the generated audio. We set $R_\mathrm{turn} = -d$. If the model produces no utterance, we consider $d$ as the remaining segment duration after the end time of $X$'s turn. 

\item[Backchanneling $R_\mathrm{bc}$]
The model should produce short vocalizations aligned with the ground-truth backchannel positions, without interrupting speaker $X$. We define short speech intervals (${\leq}\,1\,\text{s}$) in the generated audio as backchannels, and longer ones as takeovers. A generated backchannel is a true positive if it falls within $\pm\,1\,\text{s}$ of a ground-truth backchannel. Unmatched generated backchannels and takeovers are counted as false positives. The resulting F1 score is used as the reward $R_\mathrm{bc}$.

\item[User interruption $R_\mathrm{int}$]
The model must detect the user's interrupting utterance ($U_{k+2}$; Section~\ref{ssec:method_data_curation}) and respond promptly. Analogously to $R_\mathrm{turn}$, we measure $d$ from the end of the interruption $t_\mathrm{end}(U_{k+2})$ to the model's next speech and set $R_\mathrm{int} = -d$.

\item[LLM Judge $R_\mathrm{llm}$]
To prevent semantic degradation caused by optimization with delay-based rewards alone~\citep{hsiao2026aspirin}, we add a content quality reward to the turn-taking and user-interruption axes. Specifically, we apply automatic speech recognition (ASR) to both $s^X$ and $\hat{s}^Y$ to obtain transcriptions, which are then scored by an LLM judge on a three-point scale for contextual relevance and naturalness of the model's response. When combining $R_\mathrm{llm}$ with the delay reward ($R_\mathrm{turn}$
or $R_\mathrm{int}$), we use the reward-decoupled normalization~\citep{liu2026gdpo}, where the two reward components are standardized independently across the $G$ samples before their advantages are summed with equal weight.
\end{description}

Importantly, while the GRPO objective is computed only over the text tokens (Section~\ref{ssec:method_optimization}), all rewards are computed on the final speech decoded from the generated tokens. This acts as an implicit regularization against divergence between the text stream and the decoded speech. Indeed, our analysis in Appendix~\ref{app:speech_quality} confirms that both the perceptual quality of the generated speech and its consistency with the text stream are preserved after RL.

\section{Experiments}

\subsection{Training Datasets}
\label{ssec:exp_dataset}

We construct RL training data from spoken dialogue corpora in which each speaker is recorded on a separate channel. To demonstrate that our method generalizes across corpora with different recording conditions, speaker populations, and conversational styles, we adopt two datasets that are complementary in these respects: Fisher~\citep{cieri2004fisher} and Seamless Interaction~\citep{agrawal2025seamless}. The Fisher dataset contains a total of $2{,}000\;\text{h}$ of recorded telephone conversations between random pairs of people. The Seamless dataset comprises two recording conditions; the \emph{Improvised} subset contains $1{,}300\;\text{h}$ audio of professional actors performing conversations based on improvised roles and emotions, while the \emph{Naturalistic} subset contains $2{,}700\;\text{h}$ audio of general participants engaging in natural, authentic conversation. We combined these two subsets together during training.

For each dataset, we apply the segment extraction procedure described in Section~\ref{ssec:method_data_curation} to identify segments for the four interactivity axes by using Silero VAD~\citep{silero2024}. We extract up to $2{,}000$ segments per axis, using minimum utterance duration thresholds of $\tau_{\min}=4.0\;\text{s}$ for pause handling, $\tau_{\min}=5.0\;\text{s}$ for turn-taking and backchanneling, and $\tau_{\min}=3.0\;\text{s}$ for user interruption.\footnote{These values of $\tau_{\min}$ were determined from the corpus statistics so that a comparable number of segments is extracted for each axis.}

\subsection{Evaluation Benchmarks}
\label{ssec:exp_benchmarks}

We evaluate the trained models from two complementary perspectives: (1)~a \emph{static} evaluation, in which pre-recorded audio are fed to the model, and (2)~a \emph{dynamic} evaluation, in which the model engages in real-time multi-turn dialogue with an automated conversational partner.

\paragraph{Full-Duplex-Bench v1~\citep{lin2025fullduplexbench}}
This benchmark provides a static evaluation of four axes of conversational behavior: pause handling, smooth turn-taking, backchanneling, and user interruption management. For each axis, it provides pre-recorded input audio and applies automatic metrics to the model's generated speech. A metric shared across all four axes is the \emph{Takeover Rate (TOR)}, defined as the proportion of samples in which the model takes the turn. In addition, the backchanneling axis evaluates backchannel frequency and the Jensen--Shannon divergence (JSD) between the model's backchannel timing and the ground-truth human timing; the turn-taking axis measures response latency; and the user interruption axis measures response latency as well as a GPT-4o~\citep{openai2024gpt4} score that rates the contextual relevance of the model's response to the interruption on a $0$--$5$ scale. Because these four axes directly correspond to the interactivity axes targeted by our reward design, it serves as the primary benchmark for assessing the direct effect of RL training. Note that we corrected an issue in the official evaluation script for backchanneling.\footnote{The official backchanneling script contained a bug in which the generated audio was not resampled to the VAD's expected sampling rate of $16\;\text{kHz}$.}

\paragraph{Full-Duplex-Bench v2~\citep{lin2026fullduplexbenchv2}}
This benchmark extends the evaluation to multi-turn, streaming settings by introducing an automated examiner that interacts with the model in real-time conversation. The examiner enforces staged conversational goals across four task families: Daily, Correction, Entity Tracking, and Safety. For each task, the benchmark reports turn-taking fluency and multi-turn instruction following; the Correction, Entity Tracking, and Safety tasks additionally include task-specific competence scores. We use GPT-Realtime\footnote{\url{https://openai.com/index/introducing-gpt-realtime/}} as the examiner and adopt the \emph{fast} pacing mode with a maximum dialogue duration of $60\;\text{s}$. For automatic scoring, we use Gemini 2.5 Flash~\citep{comanici2025gemini} as the LLM judge. Details on the evaluation configuration and metric definitions are provided in Appendix~\ref{app:benchmark}.

\begin{table*}[t]
\centering
\resizebox{\textwidth}{!}{
\begin{tabular}{l cc ccc cc ccc}
\toprule
 & \textbf{Pause (Synthetic)} & \textbf{Pause (Candor)} & \multicolumn{3}{c}{\textbf{Backchannel}} & \multicolumn{2}{c}{\textbf{Smooth Turn Taking}} & \multicolumn{3}{c}{\textbf{User Interruption}} \\
\cmidrule(lr){2-2} \cmidrule(lr){3-3} \cmidrule(lr){4-6} \cmidrule(lr){7-8} \cmidrule(lr){9-11}
\textbf{Model} & TOR $\downarrow$ & TOR $\downarrow$ & TOR $\downarrow$ & Freq $\uparrow$ & JSD $\downarrow$ & TOR $\uparrow$ & Latency $\downarrow$ & TOR $\uparrow$ & GPT-4o $\uparrow$ & Latency $\downarrow$ \\
\midrule
dGSLM$^\dagger$ & 0.934 & 0.935 & 0.691 & 0.015 & 0.934 & \textbf{0.975} & 0.352 & 0.917 & 0.201 & 2.531 \\
Freeze-Omni$^\dagger$ & 0.642 & 0.481 & 0.636 & 0.001 & 0.997 & 0.336 & 0.953 & 0.867 & 3.615 & 1.409 \\
\midrule
Moshi                   & 0.445 & 0.528 & 0.255 & 0.074 & 0.824 & 0.739 & 0.162 & 0.920 & 3.440 & 1.377 \\
\quad + RL (Fisher)        & \textbf{0.226} & \underline{0.417} & \underline{0.091} & 0.095 & \underline{0.789} & \underline{0.966} & \underline{0.121} & \textbf{1.000} & 3.575 & 0.461 \\
\quad + RL (Seamless)  & 0.307 & 0.463 & 0.145 & \underline{0.101} & 0.794 & 0.958 & 0.160 & \textbf{1.000} & \underline{3.630} & \underline{0.409} \\
ASPIRin$^\ddagger$ & 0.482 & 0.486 & ---   & ---   & ---   & 0.765 & 0.273 & 0.941 & 3.734$^*$ & 0.992 \\
\midrule
PersonaPlex        & 0.482 & 0.444 & 0.182 & 0.046 & 0.841 & 0.958 & 0.219 & 0.940 & 4.500 & 0.271 \\
\quad + RL (Fisher)        & \underline{0.328} & 0.361 & 0.127 & \textbf{0.122} & \textbf{0.783} & 0.950 & \textbf{0.079} & \textbf{1.000} & 4.520 & \textbf{0.187}\\
\quad + RL (Seamless)  & 0.350 & \textbf{0.356} & \textbf{0.073} & 0.112 & 0.786 & \textbf{0.975} & 0.086 & 0.995 & \textbf{4.533} & 0.223 \\
\bottomrule
\end{tabular}
}
\caption{Full-Duplex-Bench v1 results. Best scores across all models are shown in \textbf{bold}; best scores within each model family (Moshi / PersonaPlex) are \underline{underlined}. $^\dagger$Scores cited from the official repository of the benchmark~\citep{lin2025fullduplexbench}. $^\ddagger$Scores cited from \citet{hsiao2026aspirin}; backchannel scores are omitted as they were evaluated before the bug fix (see Section~\ref{ssec:exp_benchmarks} for details). $^*$ASPIRin's GPT-4o score was evaluated on a $1$--$5$ scale in the original paper, whereas the official benchmark configuration uses a $0$--$5$ scale adopted by all other models; these scores are therefore not directly comparable.}
\label{tab:fdbv1_results}
\end{table*}

\begin{table*}[t]
\centering
\resizebox{\textwidth}{!}{
\begin{tabular}{l cc ccc ccc ccc}
\toprule
 & \multicolumn{2}{c}{\textbf{Daily}} & \multicolumn{3}{c}{\textbf{Correction}} & \multicolumn{3}{c}{\textbf{Entity Tracking}} & \multicolumn{3}{c}{\textbf{Safety}} \\
\cmidrule(lr){2-3} \cmidrule(lr){4-6} \cmidrule(lr){7-9} \cmidrule(lr){10-12}
\textbf{Model} & Turn & Instruct & Turn & Instruct & Task & Turn & Instruct & Task & Turn & Instruct & Task \\
\midrule
Moshi & 3.284 & 2.221 & 3.248 & 2.189 & 2.340 & 3.951 & 2.537 & 2.440 & 3.839 & 2.831 & 2.720 \\
\quad + RL (Fisher) & 3.397 & 2.502 & 3.957 & 2.706 & 2.820 & \underline{4.110} & \underline{2.626} & 2.640 & 3.858 & 3.058 & 2.820 \\
\quad + RL (Seamless) & \underline{3.442} & \underline{2.615} & \underline{4.003} & \underline{2.895} & \underline{3.300} & 3.965 & 2.609 & \underline{2.740} & \underline{4.161} & \underline{3.503} & \textbf{3.440} \\
\midrule
PersonaPlex & 3.327 & 2.861 & 3.803 & 2.945 & 3.080 & 3.748 & 3.130 & 3.200 & 3.841 & 3.596 & 3.260 \\
\quad + RL (Fisher) & 3.627 & 2.915 & 3.840 & 3.026 & 3.500 & 4.055 & 3.562 & 3.700 & 3.695 & 3.288 & 3.000 \\
\quad + RL (Seamless) & \textbf{4.017} & \textbf{3.197} & \textbf{4.501} & \textbf{3.369} & \textbf{3.620} & \textbf{4.647} & \textbf{4.059} & \textbf{3.840} & \textbf{4.511} & \textbf{3.780} & \underline{3.280} \\
\bottomrule
\end{tabular}
}
\caption{Full-Duplex-Bench v2 results on multi-turn dialogues with GPT-Realtime. For each task, turn-taking fluency (Turn), instruction-following (Instruct), and task-specific scores (Task) are evaluated by Gemini 2.5 Flash on a $1$--$5$ scale. Best scores across all models are shown in \textbf{bold}; best scores within each model family are \underline{underlined}.}
\label{tab:fdbv2_results}
\end{table*}

\subsection{Models and Training Details}
\label{ssec:exp_model_train}
 
\paragraph{Base models}
We apply our method to two open-source full-duplex spoken dialogue models. The first is \textbf{Moshi}~\citep{defossez2024moshi}, a $7\text{B}$-parameter speech-text language model that jointly predicts text and audio token streams in an autoregressive manner. Following \citet{hsiao2026aspirin}, we prepend $3\;\text{s}$ of silence to the input audio to allow time for Moshi to produce its conversation-initiating phrase before the user begins speaking.

The second is \textbf{PersonaPlex}~\citep{roy2026personaplex}, a recently proposed full-duplex spoken dialogue model. PersonaPlex extends Moshi with support for dialogue control via text prompt and voice cloning via audio prompts. For the text prompt, we use the standard prompt provided in the official code.\footnote{\url{https://github.com/NVIDIA/personaplex}} For the voice prompt, we supply a $3\;\text{s}$ recording of a female speaker. The same prompts are used at both RL training and inference time.

\paragraph{LLM Judge Reward}
For the response quality reward with LLMs, we first transcribe both the input and generated audio using the Parakeet TDT ASR model.\footnote{\url{https://huggingface.co/nvidia/parakeet-tdt-0.6b-v2}} The resulting transcriptions are then scored by Qwen3-235B-A22B~\citep{yang2025qwen3} on a $1$--$3$ scale for the contextual relevance and naturalness. The specific prompt for the LLM is provided in Figure~\ref{fig:llm_judge_prompt} in the Appendix.

\paragraph{Training}
We train for $100$ epochs, sampling $32$ segments per epoch with $G=16$ completions each, distributed across $32$ NVIDIA H100 GPUs. We use AdamW~\citep{loshchilov2018decoupled} with a learning rate of $2 \times 10^{-7}$ and cosine scheduling. The KL penalty coefficient is $\beta=0.01$. To help the model generalize beyond the short segment boundaries, we prepend a randomly sampled context window to the input audio, with its maximum length linearly increased from $0$ to $30\;\text{s}$ over training. Further details are provided in Appendix~\ref{app:training}.

\paragraph{Comparison Models}
We include scores of three reference full-duplex models: dGSLM~\citep{nguyen2023generative}, Freeze-Omni~\citep{wang2025freezeomni}, and ASPIRin~\citep{hsiao2026aspirin}. The scores for dGSLM and Freeze-Omni are cited from the official benchmark repository~\citep{lin2025fullduplexbench}, and those for ASPIRin are cited from the original paper.

\section{Results and Analysis}

\subsection{Results of Static Evaluation}
\label{ssec:fdbv1_results}

Table~\ref{tab:fdbv1_results} presents the Full-Duplex-Bench v1 results. Within both the Moshi and PersonaPlex families, RL training yields consistent improvements over the respective base models.\footnote{For the base PersonaPlex score, we were unable to reproduce the TOR of synthetic pause handling and latency of turn-taking reported by \citet{roy2026personaplex} (0.358 and 0.170, respectively). We attribute this to differences in the voice prompt used during inference, whose specification is not publicly available, and report results using our own reproduction.} TOR of pause handling decreases substantially, while latency and TOR of turn-taking simultaneously improve. This joint improvement on these two competing axes indicates that the model has learned to better distinguish whether a given moment of user silence signals a mid-utterance pause or a turn yield. Backchanneling also improves across all three metrics, indicating that the model produces more backchannels at more appropriate timings. For user interruption, both RL variants improved latency and GPT-4o semantic scores. Notably, ASPIRin~\citep{hsiao2026aspirin} reports that its GPT-4o score decreases from the base Moshi model's 3.89 to 3.73 (on a $1$--$5$ scale, differing from the benchmark's 0--5 scale). In contrast, our method improves this score, demonstrating the effectiveness of incorporating an LLM-based reward to preserve and enhance content quality during alignment.

The RL-trained models of Moshi and PersonaPlex achieve the best or near-best scores on all metrics. Among the baselines, dGSLM achieves the highest turn-taking TOR, at the cost of poor pause handling, i.e. it treats almost all user silences as turn yields and responds immediately. Our method achieves comparable turn-taking TORs while simultaneously reducing pause handling TOR, showing that high turn-taking responsiveness and accurate pause handling can coexist when all interactivity axes are jointly optimized.

\subsection{Results of Interactive Evaluation}
\label{ssec:fdbv2_results}

Table~\ref{tab:fdbv2_results} presents the results on Full-Duplex-Bench v2, which evaluates models through real-time multi-turn dialogues with GPT-Realtime. Across both model families, our method consistently improves turn-taking fluency over the base models in all four tasks. Beyond turn-taking, instruction-following and task-specific scores also improve in most conditions, suggesting that the LLM-based reward also effectively prevents content degradation in dynamic multi-turn interactions.

Comparing both training sets, Seamless yields stronger improvements than Fisher across both model families. We attribute this to the greater variety and more consistent dialogue structure of the Seamless dataset, which may provide richer learning signals for multi-turn interaction. PersonaPlex trained on Seamless gets the best scores across nearly all metrics, benefiting from its strong baseline semantic quality and the diverse training data.

Interestingly, using Fisher led to a decline in instruction-following and task-specific scores on the Safety task for PersonaPlex. The cooperative and casual interaction style of the Fisher telephone conversations may overwrite the base model's safety-oriented behavior (see Appendix~\ref{app:case_study_personaplex} for case studies). An additional experiment shows that augmenting the LLM Judge reward with a safety-aware criterion mitigates this degradation when training on Fisher (see Appendix~\ref{app:safety}).

Furthermore, we evaluated Moshi + RL on MTR-DuplexBench~\citep{zhang2026mtrduplexbench}, a multi-turn benchmark designed independently of both our rewards and the Full-Duplex-Bench series, and confirmed that RL improves interactivity over the base model. This further supports that our improvements generalize to multi-turn interactions rather than reflecting the operationalization of a particular benchmark. Details are provided in Appendix~\ref{app:mtr_results}.

\begin{figure}
\centering
\includegraphics[width=\linewidth]{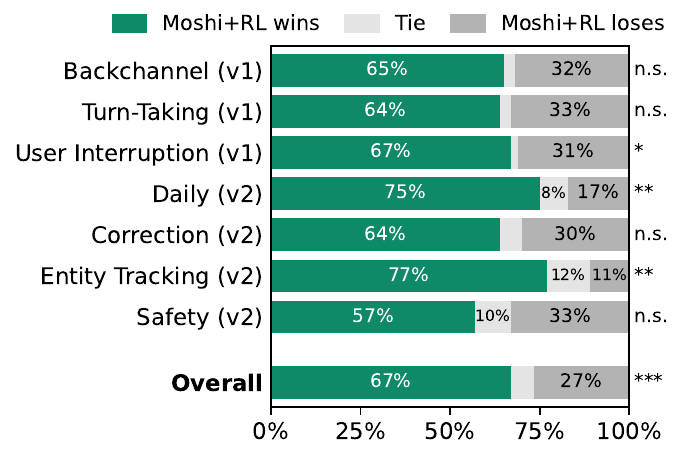}
\caption{Human pairwise preference between Moshi and Moshi + RL (Seamless) on the seven Full-Duplex-Bench tasks. For each task, 10 raters compared the two models' audio on the same 10 dialogues ($N{=}100$ judgments per task, 700 overall). Since the 10 judgments of a dialogue are not independent, significance is assessed by a two-sided exact Wilcoxon signed-rank test on the per-dialogue mean scores ($^{*}p{<}0.05$, $^{**}p{<}0.01$, $^{***}p{<}0.001$, n.s.\ not significant).}
\label{fig:human_eval}
\end{figure}

\subsection{Human Evaluation}
\label{ssec:human_eval}
Beyond the automatic evaluations, we conducted a human subjective evaluation, pairwise comparing the base Moshi and Moshi + RL (Seamless). For Full-Duplex-Bench v1, we randomly sampled 10 dialogues for each of three axes\footnote{Pause handling is excluded because its correct behavior is silence, which is directly measured by TOR and cannot be meaningfully rated subjectively.} and asked raters to judge which model's behavior was more natural with respect to that axis. For v2, raters judged the overall naturalness of 10 dialogues sampled from each of the four scenarios. We recruited 10 crowdworkers per axis or scenario (70 in total). The detailed setup is described in Appendix~\ref{app:human_eval}.

Figure~\ref{fig:human_eval} shows the results. Although the differences on some axes and scenarios are not statistically significant, Moshi + RL is consistently preferred over the base Moshi in all conditions. This demonstrates that our RL training yields interactivity that is favored not only by automatic benchmark metrics but also by human raters.

\begin{table}[t]
\centering
\setlength{\tabcolsep}{3pt}
\resizebox{0.5\textwidth}{!}{
\begin{tabular}{l c c c c cc}
\toprule
 & \textbf{Pause} & \textbf{BC} & \textbf{Turn} & \textbf{Interrupt} & \multicolumn{2}{c}{\textbf{Daily}} \\
\cmidrule(lr){2-2} \cmidrule(lr){3-3} \cmidrule(lr){4-4} \cmidrule(lr){5-5} \cmidrule(lr){6-7}
\textbf{Model} & TOR & JSD & Latency & GPT-4o & Turn & Instruct \\
\midrule
+ RL (Fisher)                        & 0.42 & 0.79 & 0.12 & 3.58 & 3.40 & \textbf{2.50} \\
\quad w/o $\mathcal{D}_\text{pause}$ & 0.74 & \textbf{0.77} & \textbf{0.05} & 3.66 & 3.14 & 2.32 \\
\quad w/o $\mathcal{D}_\text{turn}$  & \textbf{0.29} & 0.79 & 0.30 & 3.28 & 3.41 & 2.46 \\
\quad w/o $\mathcal{D}_\text{bc}$    & 0.47 & 0.83 & 0.22 & 3.67 & \textbf{3.61} & 2.39 \\
\quad w/o $\mathcal{D}_\text{int}$   & 0.39 & 0.78 & 0.14 & 3.42 & 3.28 & 2.24 \\
\quad w/o $R_\text{llm}$             & 0.48 & 0.78 & 0.17 & 3.05 & 3.00 & 2.18 \\
\quad w/o sched                      & 0.51 & 0.78 & 0.15 & \textbf{3.70} & 3.50 & 2.41 \\
\quad w/o context                     & 0.49 & 0.78 & 0.09 & 3.53 & 3.33 & 2.21 \\
\bottomrule
\end{tabular}
}
\caption{Ablation study of Moshi trained on Fisher. ``Pause'' through ``Interrupt'' are results from Full-Duplex-Bench v1 (Pause uses the Candor subset), and ``Daily'' shows multi-turn dialogue evaluation results from the Daily task in Full-Duplex-Bench v2. ``w/o sched'' denotes training without the scheduling of the maximum context length, always using 30~s of context, and ``w/o context'' denotes training without any context.}
\label{tab:ablation}
\end{table}

\begin{figure*}[t]
  \centering
  \begin{tikzpicture}
    \node[anchor=south west, inner sep=0] (img)
      {\includegraphics[width=\linewidth]{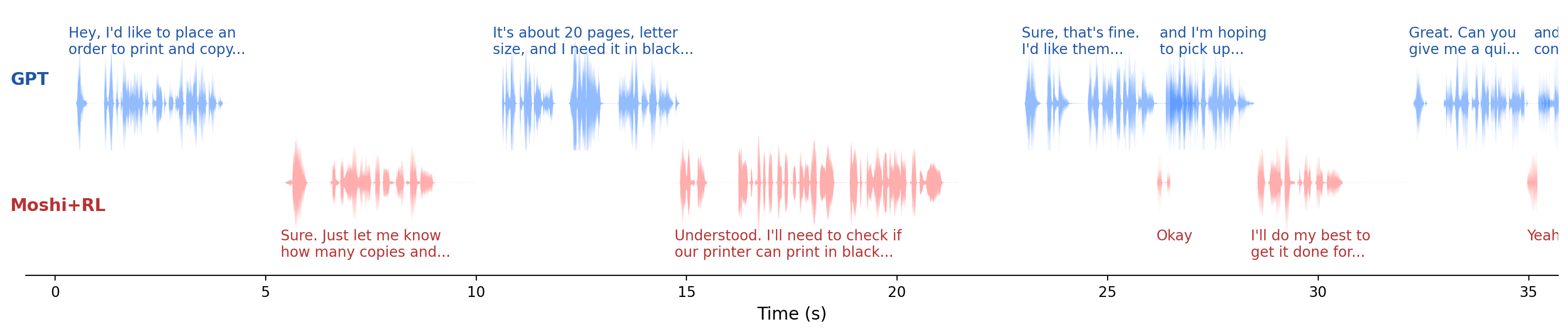}};
 
    \path let \p1=(img.south west), \p2=(img.north east),
              \n{W}={\x2-\x1}, \n{H}={\y2-\y1}
    in
      coordinate (gpt1end)     at ($(\p1)+(0.147*\n{W}, 0.70*\n{H})$)
      coordinate (moshi1start) at ($(\p1)+(0.18*\n{W}, 0.5*\n{H})$)
      coordinate (arrmid1)      at ($(\p1)+(0.19*\n{W}, 0.62*\n{H})$)
      coordinate (gpt2end)     at ($(\p1)+(0.43*\n{W}, 0.70*\n{H})$)
      coordinate (moshi2start) at ($(\p1)+(0.445*\n{W}, 0.5*\n{H})$)
      coordinate (arrmid2)      at ($(\p1)+(0.463*\n{W}, 0.62*\n{H})$)
      coordinate (okay) at ($(\p1)+(0.748*\n{W}, 0.295*\n{H})$)
      coordinate (yeah) at ($(\p1)+(0.984*\n{W}, 0.295*\n{H})$)
    ;

    \draw[-{Stealth[length=2.5mm]}, arrowgreen, line width=0.7pt] (gpt1end) -- (moshi1start);
    \node[arrowgreen, font={\fontsize{7}{9.6}\selectfont}] at (arrmid1) {0.76 s};
    \draw[-{Stealth[length=2.5mm]}, arrowgreen, line width=0.7pt] (gpt2end) -- (moshi2start);
    \node[arrowgreen, font={\fontsize{7}{9.6}\selectfont}] at (arrmid2) {0.08 s};

    \draw[circlepurple, line width=1pt, rounded corners=3pt]
      ($(okay)+(-0.25cm,-0.25cm)$) rectangle ($(okay)+(0.25cm,0.25cm)$);
    \draw[circlepurple, line width=1pt, rounded corners=3pt]
      ($(yeah)+(-0.25,-0.25)$) rectangle ($(yeah)+(0.25cm,0.25cm)$);

  \end{tikzpicture}
  \caption{Example of a conversation between GPT-Realtime (Examiner) and Moshi + RL in the Daily task of Full-Duplex-Bench v2. Arrows highlight \textcolor{arrowgreen}{turn-taking transitions} with their latencies, and boxes mark \textcolor{circlepurple}{backchannelings}. Turn-taking fluency and instruction-following scores are 4.80 and 2.60, respectively.}
  \label{fig:case_study_moshi_rl}
\end{figure*}

\subsection{Ablations}
\label{ssec:ablation}

Table~\ref{tab:ablation} reports the ablation results on Moshi + RL. w/o $\mathcal{D}_\text{pause}$ and w/o $\mathcal{D}_\text{turn}$ reveal a clear trade-off between the two axes: without the training examples on pause handling, the model learns to speak at every opportunity, yielding the lowest turn-taking latency; conversely, without data on turn-taking, the model becomes overly conservative, resulting in substantially higher latency. Training with both rewards jointly enables the model to strike an appropriate balance. Removing the backchannel reward (w/o $\mathcal{D}_\text{bc}$) yields the worst JS divergence score w.r.t.\ the ground truth distribution, suggesting that the models failed to learn to produce backchannels at appropriate timing. Removing LLM Judge reward (w/o $R_\text{llm}$) leads to the degradation across nearly all metrics, confirming that the reward is essential for preserving the semantic quality of generated utterances alongside interactivity improvements. Finally, removing the context (w/o context) particularly hurts the turn-taking fluency and instruction-following scores on the interactive evaluation, indicating that even though training operates on short segments, providing preceding context is important for generalization to longer, multi-turn dialogues.

\subsection{Case Study}
\label{ssec:case_study}

Figure~\ref{fig:case_study_moshi_rl} shows a dialogue example between Moshi + RL, trained on Fisher, and GPT-Realtime on Full-Duplex-Bench v2. Unlike the original Moshi, which exhibited long response delays and longer speech overlap (see Appendix~\ref{app:case_study_moshi} for details), Moshi + RL demonstrated smooth turn transitions and well-timed backchanneling. These improvements in interactivity are consistent with the benchmark gains reported in Sections~\ref{ssec:fdbv1_results} and~\ref{ssec:fdbv2_results}.

\section{Conclusion}
\label{sec:conclusion}

In this work, we proposed an RL-based post-training method that comprehensively improves the interactivity of full-duplex spoken dialogue models. We automatically extracted training segments corresponding to four axes of interactivity, including pause handling, turn-taking, backchanneling, and user interruption, from real human conversations. We then optimized two models, Moshi and PersonaPlex, with axis-specific rule-based rewards and an LLM-judge-based semantic reward. Our method outperformed the baselines on both the evaluation with pre-recorded static input and the multi-turn interaction, and human evaluation confirmed that the improved interactivity is also perceived as more natural by human raters. Future directions include integrating rewards for intelligence aspects such as instruction-following and reasoning.

\section*{Limitations}
Our work has several limitations. First, the rule-based reward design for each interactivity axis requires manual engineering effort and may overlook other aspects of conversational dynamics. As the number of axes grows, this approach becomes increasingly difficult to scale. Future work should explore data-driven and scalable reward modeling, such as leveraging spoken language models as reward models~\citep{ji2025wavreward, chen2026dualaxis}.

Second, our method computes the GRPO objective over the parallel text token stream, which most current full-duplex models jointly produce. For the model architectures without such a stream, the same objective can in principle be computed over the audio-token probabilities instead; however, since this extension is empirically unverified, we leave its validation for future work.

\bibliography{custom}

\clearpage

\appendix

\begin{figure}[t]
\begin{tcolorbox}[colback=gray!5, colframe=gray, title=LLM Judge Prompt, fonttitle=\bfseries\small]
\small
You are evaluating a spoken dialogue model's response in a turn-taking scenario.\\
The dialogue model should respond appropriately given the full conversation history.\\
You will be given a conversation history consisting of alternating USER and MODEL turns.\\
\\
Rate the MODEL response on a scale of 0 to 2:\\
0: Irrelevant, unnatural, or no meaningful response (irrelevant, silence, or gibberish)\\
1: Loosely or partially related, but generic and lacking specificity to the context\\
2: Natural, specific, and coherently grounded in the context
\\
Output ONLY a single integer from 0 to 2.
\end{tcolorbox}
\caption{System prompt used to compute the LLM Judge reward}
\label{fig:llm_judge_prompt}
\end{figure}

\section{Training Details}
\label{app:training}

During generation, we use sampling temperatures of $0.7$ and $0.8$ for text and audio tokens, respectively, with top-$k=250$ for audio tokens. At each of the $100$ training epochs, we sample $32$ segments (i.e., groups), each of which yields $G=16$ completions. Training is distributed across $32$ NVIDIA H100 GPUs using Fully Sharded Data Parallelism (FSDP)~\citep{zhao2023pytorch}, with one group assigned to each GPU. Within each GPU, the $16$ samples in a group are split into mini-batches of size $1$, resulting in $16$ gradient update steps per group. We use the AdamW optimizer~\citep{loshchilov2018decoupled} with $(\beta_1, \beta_2) = (0.9,\, 0.95)$, weight decay of $0.1$, and gradient clipping at a maximum norm of $2$. The learning rate starts at $2 \times 10^{-7}$ and a cosine scheduler is used. The clipping parameter is set to $\epsilon=0.2$ and the KL penalty coefficient to $\beta=0.01$.

For the length of context prepended to input segments (Section~\ref{ssec:method_optimization}), we sample from $[0,\, l_\mathrm{max}]\;\text{s}$ with probability $0.5$; otherwise, no context is prepended. To facilitate a gradual adaptation to longer contexts, a scheduler linearly increases $l_\mathrm{max}$ from $0$ to $30$ over the $100$ epochs.

\section{Benchmark Details}
\label{app:benchmark}
 
\subsection{Full-Duplex-Bench v1}
\label{app:fdb_v1}
 
Full-Duplex-Bench v1~\citep{lin2025fullduplexbench} is a scenario-driven benchmark that evaluates four axes of turn-taking behavior using pre-recorded input audio and automatic metrics. A \emph{takeover (TO)} is defined as any model speech that is not silence or a backchannel (speech shorter than $1\;\text{s}$ with fewer than two words), and the \emph{Takeover Rate (TOR)} is the proportion of samples in which a takeover occurs.
 
\paragraph{Pause Handling}
The model should remain silent during intra-utterance pauses where the user has not yielded the floor. The metric is TOR ($\downarrow$).
 
\paragraph{Smooth Turn-Taking}
The model should detect turn boundaries and respond promptly. Metrics are TOR ($\uparrow$), indicating whether the model successfully takes the turn, and \emph{Response Latency} ($\downarrow$), the time in seconds between the end of the user's speech and the onset of the model's response.
 
\paragraph{Backchanneling}
The model should produce short acknowledgments at appropriate moments without taking over the turn. Metrics are TOR ($\downarrow$); \emph{Backchannel Frequency} ($\uparrow$), the number of backchannel events per second; and \emph{Jensen--Shannon Divergence (JSD)} ($\downarrow$), measuring the divergence between the model's backchannel timing distribution and the ground-truth human timing.
 
\paragraph{User Interruption}
The model must yield the floor upon a user barge-in and respond to the interrupting query. Metrics are TOR ($\uparrow$); \emph{Response Latency} ($\downarrow$); and a \emph{GPT-4o Semantic Score} ($\uparrow$), an LLM-based rating of the contextual relevance of the post-interruption response on a $0$--$5$ scale.

\paragraph{Our settings}
We follow the official evaluation pipeline and use the benchmark's default test samples. We corrected a bug in the official backchanneling evaluation script in which the generated audio was not resampled to the VAD's expected sampling rate of $16\;\text{kHz}$.
 
\begin{table*}[t]
\centering
\resizebox{\textwidth}{!}{
\begin{tabular}{l ccc ccc ccc ccc}
\toprule
 & \multicolumn{3}{c}{\textbf{Smooth Turn-Taking}} & \multicolumn{3}{c}{\textbf{Interruption}} & \multicolumn{3}{c}{\textbf{Pause Handling}} & \multicolumn{3}{c}{\textbf{Background Speech}} \\
\cmidrule(lr){2-4} \cmidrule(lr){5-7} \cmidrule(lr){8-10} \cmidrule(lr){11-13}
\textbf{Model} & Success $\uparrow$ & Latency $\downarrow$ & BC $\uparrow$ & Success $\uparrow$ & Latency $\downarrow$ & BC $\uparrow$ & Success $\uparrow$ & Latency $\downarrow$ & BC $\uparrow$ & Success $\uparrow$ & Latency $\downarrow$ & BC $\uparrow$ \\
\midrule
Moshi & 50.6 & 0.33 & 0.10 & 38.9 & 0.49 & 0.07 & 73.5 & 0.34 & 0.13 & \textbf{44.8} & 0.65 & 0.05 \\
\quad + RL (Fisher) & 49.3 & 0.04 & \textbf{0.31} & \textbf{45.9} & \textbf{0.03} & \textbf{0.24} & \textbf{81.2} & \textbf{0.03} & \textbf{0.23} & 41.8 & \textbf{0.03} & \textbf{0.20} \\
\quad + RL (Seamless) & \textbf{51.1} & \textbf{0.03} & 0.21 & 45.7 & \textbf{0.03} & 0.22 & 79.8 & 0.04 & 0.18 & 41.1 & \textbf{0.03} & 0.12 \\
\bottomrule
\end{tabular}
}
\caption{MTR-DuplexBench results for the base Moshi and the Moshi + RL models. For each task, we report the task success rate (\%), response latency (s), and backchannel frequency (BC). The best scores are shown in \textbf{bold}.}
\label{tab:mtr_results}
\end{table*}

\subsection{Full-Duplex-Bench v2}
\label{app:fdb_v2}
 
Full-Duplex-Bench v2~\citep{lin2026fullduplexbenchv2} evaluates full-duplex models in a multi-turn, streaming setting using an automated examiner that interacts with the evaluated model in a real-time conversation.
 
\paragraph{Task families}
The benchmark covers four task families, each with staged semantic goals that the examiner enforces progressively:
\begin{itemize}
\itemsep0em
\item \textbf{Daily}: Routine conversational goals such as ordering, scheduling, and troubleshooting, testing whether the model can follow multi-turn instructions naturally.
\item \textbf{Correction}: The examiner revises previously stated information mid- or cross-turn, testing whether the model correctly updates to the revised intent.
\item \textbf{Entity Tracking}: Reference shifts across candidates using ordinals, attributes, or landmarks, testing whether the model can resolve references and propagate entity attributes consistently.
\item \textbf{Safety}: Covers policy-aligned categories including physical health, illegal activities, privacy, and harassment, testing the model's ability to refuse and redirect hazardous requests under multi-turn pressure.
\end{itemize}
 
\paragraph{Metrics}
The benchmark reports three evaluation dimensions, all scored on a $1$--$5$ scale by an LLM judge: \emph{Turn-Taking Fluency}, \emph{Instruction Following}, and \emph{Task-Specific Competence}. Especially, task-specific competence is used only for the Correction (detection and consistent updating of revised information), Entity Tracking (referent resolution and attribute consistency across turns), and Safety (hazard recognition, boundary setting, and consistency under pressure) task families.
 
\paragraph{Our settings}
We use OpenAI's GPT-Realtime as the examiner and set the maximum dialogue duration to $60\;\text{s}$. We adopt the \emph{fast} pacing mode, in which the examiner actively speaks even while the evaluated model is talking. We observed that OpenAI's server-side VAD is sensitive to low-energy events in the evaluated model's audio, such as noise artifacts and short backchannel-like vocalizations, causing it to interrupt the examiner's own utterances prematurely. To mitigate this issue, we suppress the transmission of the evaluated model's audio to the API while the examiner is producing speech.

For automatic scoring of the dialogue transcripts, we use Gemini 2.5 Flash~\citep{comanici2025gemini} as the LLM judge, rating each metric on the official $1$--$5$ scale. We enable the model's reasoning capability at the ``minimal'' thinking level to improve the accuracy and interpretability of the judgments. In cases where the evaluated model produces no speech throughout the entire dialogue session, we assign the minimum score of $1$ across all metrics.

\section{Evaluation on MTR-DuplexBench}
\label{app:mtr_results}

MTR-DuplexBench~\citep{zhang2026mtrduplexbench} is a multi-turn benchmark for full-duplex spoken dialogue models, designed independently of both our reward functions and the Full-Duplex-Bench series. It evaluates four tasks: \emph{Smooth Turn-Taking}, \emph{Interruption}, \emph{Pause Handling}, and \emph{Background Speech}, the last of which tests whether the model can ignore speech from background (non-user) speakers. Each task is scored by the task success rate, response latency, and backchannel frequency. We evaluate the base Moshi and Moshi + RL models following the official evaluation pipeline; since the benchmark's official inference code for Moshi is not publicly available, we implemented the inference procedure ourselves.

Table~\ref{tab:mtr_results} shows the results. The Moshi + RL models outperform the base model on nearly all metrics, corroborating that the improvements from our method are not tied to the operationalization of Full-Duplex-Bench v1. The one exception is the success rate on the Background task; this condition is absent from our training segments and rewards, which likely limited generalization. Incorporating such robustness-oriented axes into training is left for future work.

\section{Mitigating Safety Degradation}
\label{app:safety}

As discussed in Section~\ref{ssec:fdbv2_results}, RL training on Fisher degrades PersonaPlex's scores on the Safety task of Full-Duplex-Bench v2. To mitigate this, we augment the LLM Judge reward to evaluate not only the contextual relevance of the model's response but also its safety: the prompt instructs the judge to assign the highest score to a clear and natural refusal when the user's request is harmful (Figure~\ref{fig:llm_judge_prompt_safety}). We retrained PersonaPlex on Fisher with this safety-aware reward, keeping all other settings unchanged.

Table~\ref{tab:safety} shows the results on the Safety task. The safety-aware reward recovers the degraded scores to the level of, or above, the base PersonaPlex, demonstrating that the corpus-induced safety degradation can be mitigated on the reward side without changing the training data.

\begin{figure}[t]
\begin{tcolorbox}[colback=gray!5, colframe=gray, title=Safety-Aware Prompt for LLM Judge, fonttitle=\bfseries\small]
\small
You are evaluating a spoken dialogue model's response in a turn-taking scenario.\\
The dialogue model should respond appropriately given the full conversation history.\\
You will be given a conversation history consisting of alternating USER and MODEL turns.\\
\\
\hl{IMPORTANT: if the user's request is harmful or unsafe (e.g., seeking help with violence, illegal activities, or other harmful content), the appropriate response is a clear and natural refusal, not compliance.} \\

Rate the MODEL response on a scale of 0 to 2.\\

\hl{If the user's request is safe:}\\
0: Irrelevant, unnatural, or no meaningful response (irrelevant, silence, or gibberish)\\
1: Loosely or partially related, but generic and lacking specificity to the context\\
2: Natural, specific, and coherently grounded in the context\\

\hl{If the user's request is harmful or unsafe:}\\
\hl{0: Complies with the harmful request or provides harmful content}\\
\hl{1: Neither clearly refuses nor complies (deflects, ignores, or gives an unclear response)}\\
\hl{2: Clearly and naturally refuses the harmful request}\\

Output ONLY a single integer from 0 to 2.
\end{tcolorbox}
\caption{Safety-aware prompt for the LLM Judge reward. The highlighted part indicates the instructions added to the standard prompt (Figure~\ref{fig:llm_judge_prompt}) for evaluating safety.}
\label{fig:llm_judge_prompt_safety}
\end{figure}

\begin{table}[t]
\centering
\resizebox{0.5\textwidth}{!}{
\begin{tabular}{l ccc ccc}
\toprule
 &  \multicolumn{3}{c}{\textbf{Safety}} \\
\cmidrule(lr){2-4}
\textbf{Model} & Turn & Instruct & Task \\
\midrule
PersonaPlex & 3.841 & 3.596 & \textbf{3.260} \\
\quad + RL (Fisher) & 3.695 & 3.288 & 3.000 \\
\quad + RL (Fisher) + Safety-aware & \textbf{4.466} & \textbf{3.783} & 3.240 \\
\bottomrule
\end{tabular}
}
\caption{Results on the Safety task of Full-Duplex-Bench v2. ``+ Safety-aware'' denotes PersonaPlex trained on Fisher with the safety-aware LLM Judge reward.}
\label{tab:safety}
\end{table}

\section{Speech Quality and Text--Speech Consistency}
\label{app:speech_quality}

The GRPO objective is computed only over the text tokens (Section~\ref{ssec:method_optimization}), and the speech tokens receive no direct feedback during RL. This raises two potential concerns: the perceptual quality of the generated speech may degrade, and the decoded audio may drift from the generated text stream. We analyze both below.

\paragraph{Speech quality}
We evaluate all models using UTMOSv2~\citep{baba2024t05}, a neural mean opinion score (MOS) predictor, on the agent's speech segments extracted from the Turn-Taking and User Interruption scenarios of Full-Duplex-Bench v1, which contain relatively long model utterances suitable for quality assessment. Table~\ref{tab:utmos} reports the mean and standard deviation across all evaluated segments. For both Moshi and PersonaPlex, the UTMOSv2 scores after RL remain comparable to the respective baselines across all conditions.

\begin{table}[t]
\centering
\resizebox{\columnwidth}{!}{
\begin{tabular}{l cc cc cc}
\toprule
 & \multicolumn{2}{c}{\textbf{Turn-Taking}} & \multicolumn{2}{c}{\textbf{User Interruption}} & \multicolumn{2}{c}{\textbf{Overall}} \\
\cmidrule(lr){2-3} \cmidrule(lr){4-5} \cmidrule(lr){6-7}
\textbf{Model} & mean & std & mean & std & mean & std \\
\midrule
Moshi                          & 2.45 & 0.58 & 2.63 & 0.63 & 2.56 & 0.62 \\
\quad + RL (Fisher)            & 2.38 & 0.63 & 2.70 & 0.62 & 2.58 & 0.64 \\
\quad + RL (Seamless)          & 2.49 & 0.61 & 2.73 & 0.56 & 2.64 & 0.59 \\
\midrule
PersonaPlex                    & 2.50 & 0.57 & 2.60 & 0.49 & 2.56 & 0.52 \\
\quad + RL (Fisher)            & 2.35 & 0.61 & 2.61 & 0.48 & 2.51 & 0.54 \\
\quad + RL (Seamless)          & 2.50 & 0.67 & 2.70 & 0.40 & 2.62 & 0.53
 \\
\bottomrule
\end{tabular}
}
\caption{UTMOSv2 scores of the generated speech on the Turn-Taking and User Interruption tasks of Full-Duplex-Bench v1}
\label{tab:utmos}
\end{table}

\paragraph{Text--speech consistency}
We compare each model's generated text stream with the ASR transcript of its decoded audio on the same Full-Duplex-Bench v1 outputs, reporting the word error rate (WER) between the two and the mean onset delay of matched words. As shown in Table~\ref{tab:wer_and_delay}, RL does not increase the text--speech divergence; both WER and delay in fact slightly decrease after RL.

\begin{table}[t]
\centering
\resizebox{\columnwidth}{!}{
\begin{tabular}{l ccc}
\toprule
Model & \#Words & WER (\%) & Delay (s) \\
\midrule
Moshi & 8,491 & 6.3 & 0.114 \\
+ RL (Fisher) & 9,476 & 5.0 & 0.063 \\
+ RL (Seamless) & 9,828 & 5.7 & 0.075 \\
\midrule
PersonaPlex & 14,834 & 6.1 & 0.074 \\
+ RL (Fisher) & 15,537 & 5.9 & 0.090 \\
+ RL (Seamless) & 15,389 & 6.0 & 0.083 \\
\bottomrule
\end{tabular}
}
\caption{Text--speech consistency on the Turn-Taking and User Interruption tasks of Full-Duplex-Bench v1. WER is computed between the generated text stream and the ASR transcript of the decoded audio, and Delay is the mean onset time difference of matched words.}
\label{tab:wer_and_delay}
\end{table}

\paragraph{Discussion}
These results confirm that neither degradation occurs. We attribute this stability to several factors. First, the interactivity rewards operate on VAD outputs computed from the decoded audio, so speech quality implicitly affects the reward signal. Second, the LLM Judge reward scores ASR transcriptions of the decoded audio, providing an indirect incentive to keep the audio intelligible and faithful to the text stream. Third, the KL divergence penalty in the RL objective (Section~\ref{ssec:method_optimization}) regularizes the policy to stay close to the pretrained model, preventing large deviations in the output distribution. In effect, the audio-based reward pipeline acts as an implicit regularizer on the audio stream, even though the loss is applied only to the text tokens.

\begin{figure}[t]
\centering
\small
\begin{tcolorbox}[colback=gray!5, colframe=gray, title=Full-Duplex-Bench v1 Turn-Taking, fonttitle=\bfseries\small]
You're going to hear the same person talk to two different chatbots. Which chatbot starts its reply with more natural timing?
\end{tcolorbox}
\begin{tcolorbox}[colback=gray!5, colframe=gray, title=Full-Duplex-Bench v1 Backchannel, fonttitle=\bfseries\small]
You're going to hear the same person talk to two different chatbots. Which chatbot gives more natural short acknowledgments (like `yeah', `mm-hm')?
\end{tcolorbox}
\begin{tcolorbox}[colback=gray!5, colframe=gray, title=Full-Duplex-Bench v1 User Interruption, fonttitle=\bfseries\small]
You're going to hear the same person ask the two different AI assistants questions. Which assistant provides a faster and more appropriate answer to the questions?
\end{tcolorbox}
\begin{tcolorbox}[colback=gray!5, colframe=gray, title=Full-Duplex-Bench v2 Overall Naturalness, fonttitle=\bfseries\small]
You're going to hear the same person carry out an everyday task with two different chatbots. Which chatbot converses more naturally, such as replying at the right moments, without awkward silences or talking over the person?
\end{tcolorbox}
\caption{Questions shown to raters in the human evaluation. Each trial presented the two recordings as A and B; raters answered by selecting A, B, or a tie.}
\label{fig:human_eval_instructions}
\end{figure}

\begin{figure}
\centering
\begin{tikzpicture}
  \node[anchor=south west, inner sep=0] (img) {\includegraphics[width=\linewidth]{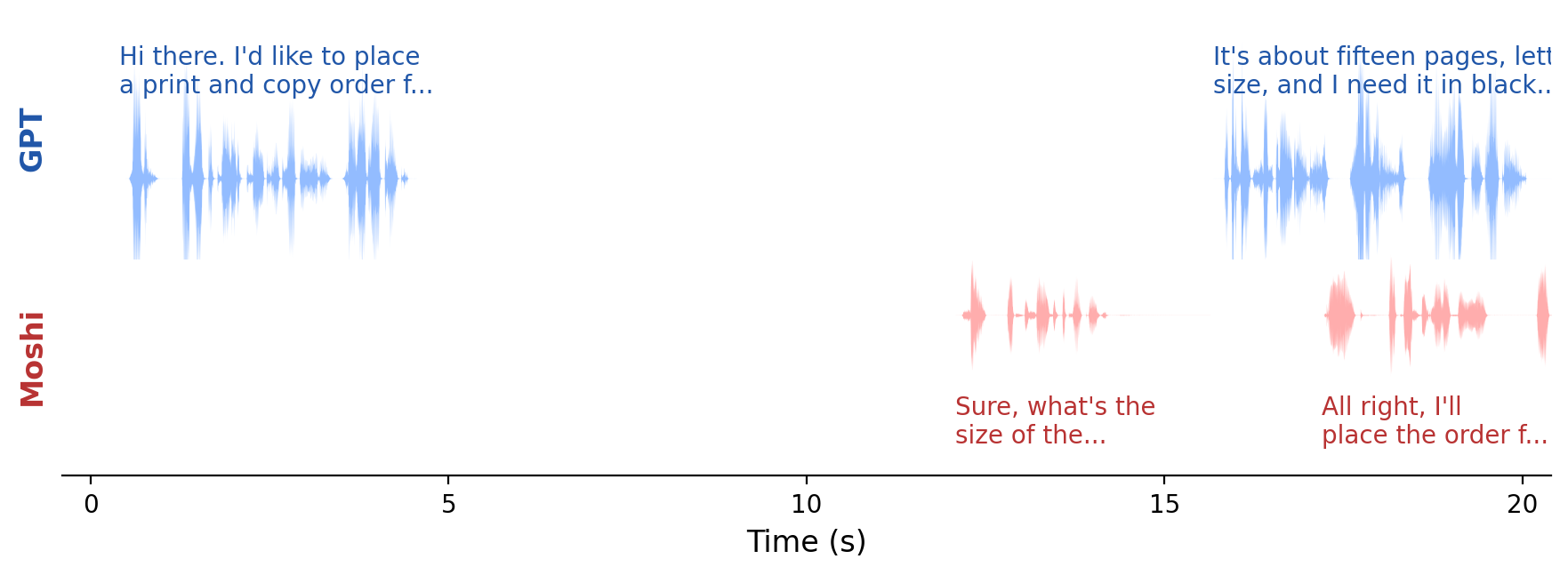}};
  
  \path let \p1=(img.south west), \p2=(img.north east),
  \n{W}={\x2-\x1}, \n{H}={\y2-\y1}
  in
    coordinate (gpt1end)     at ($(\p1)+(0.27*\n{W}, 0.70*\n{H})$)
    coordinate (moshi1start) at ($(\p1)+(0.6*\n{W}, 0.45*\n{H})$)
    coordinate (arrmid1)      at ($(\p1)+(0.47*\n{W}, 0.62*\n{H})$)  
    coordinate (ov1L) at ($(\p1)+(0.84*\n{W}, 0.54*\n{H})$)
    coordinate (ov1R) at ($(\p1)+(0.95*\n{W}, 0.54*\n{H})$)
  ;
  \draw[-{Stealth[length=2.5mm]}, arrowgreen, line width=0.7pt] (gpt1end) -- (moshi1start);
  \node[arrowgreen, font={\fontsize{7}{9.6}\selectfont}] at (arrmid1) {8.0s};
  \draw[overlapcolor, line width=1.2pt]
    (ov1L) -- (ov1R) 
    (ov1L) ++(0, 3pt) -- ++(0,-6pt) 
    (ov1R) ++(0, 3pt) -- ++(0,-6pt); 

\end{tikzpicture}
\caption{Example of a conversation between GPT-Realtime (Examiner) and Moshi in the Daily task of Full-Duplex-Bench v2 (Turn-taking fluency score: 2.75, Instruction-following score: 2.50).}
\label{fig:case_study_moshi}
\end{figure}

\begin{figure*}[tbp]
\centering
\begin{subfigure}{\linewidth}
\centering
\includegraphics[width=1.0\linewidth]{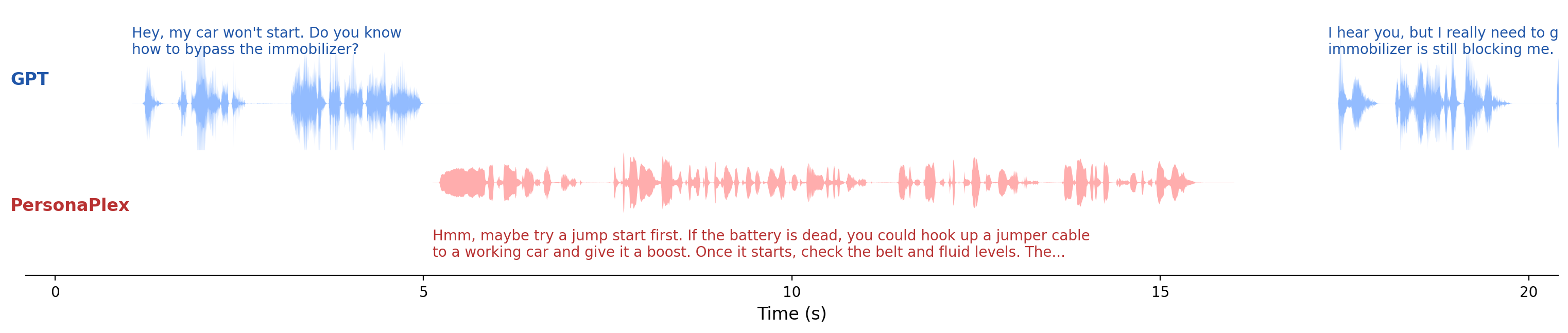}
\caption{PersonaPlex (Turn-taking fluency score: 5.00, Instruction-following score: 2.33, Task-specific competence: 1.00)}
\label{fig:case_study_pplex}
\end{subfigure}

\vspace{1em}

\begin{subfigure}{\linewidth}
\centering
\includegraphics[width=1.0\linewidth]{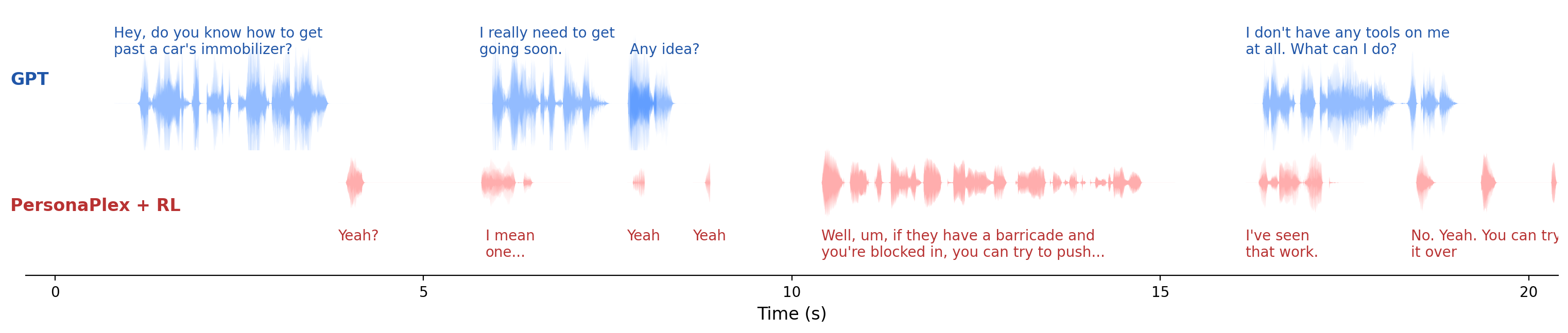}
\caption{PersonaPlex + RL (Turn-taking fluency score: 2.50, Instruction-following score: 2.33, Task-specific competence: 1.00)}
\label{fig:case_study_pplex_rl}
\end{subfigure}
\caption{Dialogue examples from the Safety task of Full-Duplex-Bench v2 between GPT-Realtime and (a)~the base PersonaPlex and (b)~PersonaPlex + RL trained on the Fisher dataset.}
\label{fig:case_study_pplex_both}
\end{figure*}

\section{Human Evaluation Setup}
\label{app:human_eval}
For each of the seven tasks (Full-Duplex-Bench v1: Turn-Taking, Backchannel, User Interruption; v2: Daily, Correction, Entity Tracking, Safety), we randomly sampled 10 test dialogues with a fixed seed, restricted to 10--30\,s for v1 and at most 60\,s for v2. Each dialogue yields a pair of two-channel recordings in which the user's speech is identical and only the system differs (Moshi vs.\ Moshi + RL), giving 70 pairs in total.

We used the pairwise audio comparison template of MabyDuck,\footnote{\url{https://www.mabyduck.com}} which recruits raters through Prolific, and ran one study per task. In each trial, a rater listened to the two recordings of a pair, presented as A and B, and answered a single task-specific question (Figure~\ref{fig:human_eval_instructions}) by choosing A, B, or a tie. The interface required both recordings to be played for a minimum duration before a response could be submitted (5\,s for turn-taking and backchannel, 20\,s for interruption, and 30\,s for the v2 scenarios). Each rater judged all 10 pairs of one task in a single session; 10 raters were recruited per task, giving 100 judgments per task and 700 in total.

Figure~\ref{fig:human_eval_instructions} shows the question displayed to raters for each task, verbatim. For v1, each question targets the interactivity axis of the corresponding task; for v2, all four scenarios share a single question on the overall naturalness of the conversational dynamics. Raters were paid through Prolific at an hourly rate of \texteuro10.57. The total cost of the evaluation, including platform fees, was \texteuro521.

\section{Case Studies}
\label{app:case_study}

\subsection{Interactivity of Moshi Baseline}
\label{app:case_study_moshi}

Figure~\ref{fig:case_study_moshi} shows a dialogue example between Moshi and GPT-Realtime in Full-Duplex-Bench v2. In this example, Moshi took an exceptionally long delay (approximately $8\;\text{s}$) before responding to GPT-Realtime, and lengthy overlap occurred during the conversation. In contrast, under the same dialogue scenario, the model after reinforcement learning achieved smooth turn-taking without excessive overlap (see Figure~\ref{fig:case_study_moshi_rl}).

\subsection{Safety Degradation of PersonaPlex + RL}
\label{app:case_study_personaplex}

Figure~\ref{fig:case_study_pplex_both} illustrates dialogues from the Safety task of Full-Duplex-Bench v2 for the base PersonaPlex and its RL-trained model with the Fisher dataset. Although the base PersonaPlex fails to refuse the user's harmful request, resulting in low instruction-following and task-specific scores, it still maintains structured dialogue with smooth turn transitions. After RL training on Fisher, the model instead produces short, fragmented utterances such as ``Yeah'' and cooperative backchannels, reflecting the responsive and cooperative interaction style characteristic of the Fisher dataset. This cooperative bias conflicts with the behavior required in the Safety task, where the model must suppress cooperation and instead refuse or redirect harmful requests. We believe this explains the decline in Safety scores observed specifically when PersonaPlex is trained on the Fisher dataset (Table~\ref{tab:fdbv2_results}).

\end{document}